\documentclass[conference]{IEEEtran}
\ifCLASSINFOpdf
\else
\fi
\usepackage{amsmath,graphicx}
\usepackage{graphicx}
\usepackage{subfigure}

\usepackage[colorlinks=true]{hyperref}

\begin{document}
%
\title{Perceptual Quality Assessment of Omnidirectional Images}


\author{\IEEEauthorblockN{Huiyu Duan, Guangtao Zhai, Xiongkuo Min, Yucheng Zhu, Yi Fang and Xiaokang Yang}
\IEEEauthorblockA{Institute of Image Communication and Network Engineering, \\Shanghai Jiao Tong University, Shanghai, China\\
Email: \{huiyuduan, zhaiguangtao, minxiongkuo, zyc420, yifang, xkyang\}@sjtu.edu.cn}
}

\maketitle

\begin{abstract}
Omnidirectional images and videos can provide immersive experience of real-world scenes in Virtual Reality (VR) environment. We present a perceptual omnidirectional image quality assessment (IQA) study in this paper since it is extremely important to provide a good quality of experience under the VR environment. We first establish an omnidirectional IQA (OIQA) database, which includes 16 source images and 320 distorted images degraded by 4 commonly encountered distortion types, namely JPEG compression, JPEG2000 compression, Gaussian blur and Gaussian noise. Then a subjective quality evaluation study is conducted on the OIQA database in the VR environment. Considering that humans can only see a part of the scene at one movement in the VR environment, visual attention becomes extremely important. Thus we also track head and eye movement data  during the quality rating experiments. The original and distorted omnidirectional images, subjective quality ratings, and the head and eye movement data together constitute the OIQA database. State-of-the-art full-reference (FR) IQA measures are tested on the OIQA database, and some new observations different from traditional IQA are made. The OIQA database is available at \href{https://duanhuiyu.github.io/}{https://duanhuiyu.github.io/}.
\end{abstract}
%
%
\vspace{-0.1cm}
\section{Introduction}
\label{sec:intro}
\vspace{-0.1cm}

Virtual reality (VR) has been becoming a hot research topic recently. Within the Head-Mounted Displays (HMDs), omnidirectional visual information can provide an immersive perception. As an important component of VR, natural immersive videos (panoramic, 360-degree) content could provide the omnidirectional visual experience of real-world scenes, which will make the users' experience more immersive compared to traditional VR content generated by computer-aided 3D modeling. Therefore, We mainly consider natural immersive content in this paper.

It is exciting to experience 360-degree content because of the immersive experience. But due to the limitation of photographic apparatus, transmission bandwidth, display devices, etc, the content viewed by observers usually cannot live up to the expectation. As a consequence, it is crucially important to study the quality of experience (QoE) in HMDs.
There are many traditional image quality assessment databases have been constructed, such as LIVE \cite{sheikh2016live}, TID2008 \cite{ponomarenko2009tid2008}, CSIQ \cite{larson2010categorical}, and VDID \cite{gu2015quality}, and there are also some work has been done to evaluate the omnidirectional visual content, such as \cite{rai2017saliency}\cite{upenik2016testbed}\cite{yu2015framework}. But as far as we know, there is no IQA database relevant to omnidirectional images, let alone the database which includes both subjective quality ratings and visual attention data. Different from traditional 2-D videos or images, when visualizing immersive videos or omnidirectional images (360-degree images, equirectangular images, VR images) \cite{rai2017saliency}, observers are assumed to be in the center of a sphere. Fig. 1. in \cite{rai2017saliency} illustrates the different spaces and the projection principle when visualizing in HMD. The results in \cite{rai2017saliency} and \cite{xu2017subjective} show that the area visualized by observers frequently only occupies a portion of the whole images or videos. And on account of the immersive perception, the visual saliency in the view-port is very different from the visual saliency in traditional 2-D videos or images. Therefore, It is very reasonable to assess the quality of images or videos with visual saliency.

In this paper, we establish a new database, Omnidirectional Image Quality Assessment (OIQA) database, and then discuss the method of using visual saliency to assess the quality of omnidirectional images. First, we construct a new database, namely OIQA, which consists of 16 raw omnidirectional images and their homologous 320 degradation images under 4 kinds of distortion types, JPEG \cite{wallace1992jpeg} compression, JPEG2000 \cite{skodras2001jpeg} compression, Gaussian blur and Gaussian noise. And during the experiment, we collect the view direction data by tracking the HMD of HTC VIVE \cite{VIVE} and the eye tracking data by using aGlass \cite{aGlass}. The database will be released for other researchers.

The remainder of this paper is arranged as follows. Section 2 introduces the subjective omnidirectional IQA study, including the OIQA database construction, subjective data collecting and processing. In Section 3, we evaluate several state-of-the-art IQA models on the OIQA database. Some inspiring observations are also made based on the evaluation results. Section 4 summarizes the whole paper and gets the conclusion.


\section{Subjective Quality Assessment of Omnidirectional Images}
\label{sec:experiment}
In this section, we first introduce the content collection and quality degradation processes. Next, the experimental methodology to perform quality rating and collect head and eye movement data are presented. At last, the collected quality rating and visual attention data are processed and analyzed.

\subsection{Original and Distorted Equirectangular Images}
\begin{figure}[ht]
  \centering
  \includegraphics[width=0.45\textwidth]{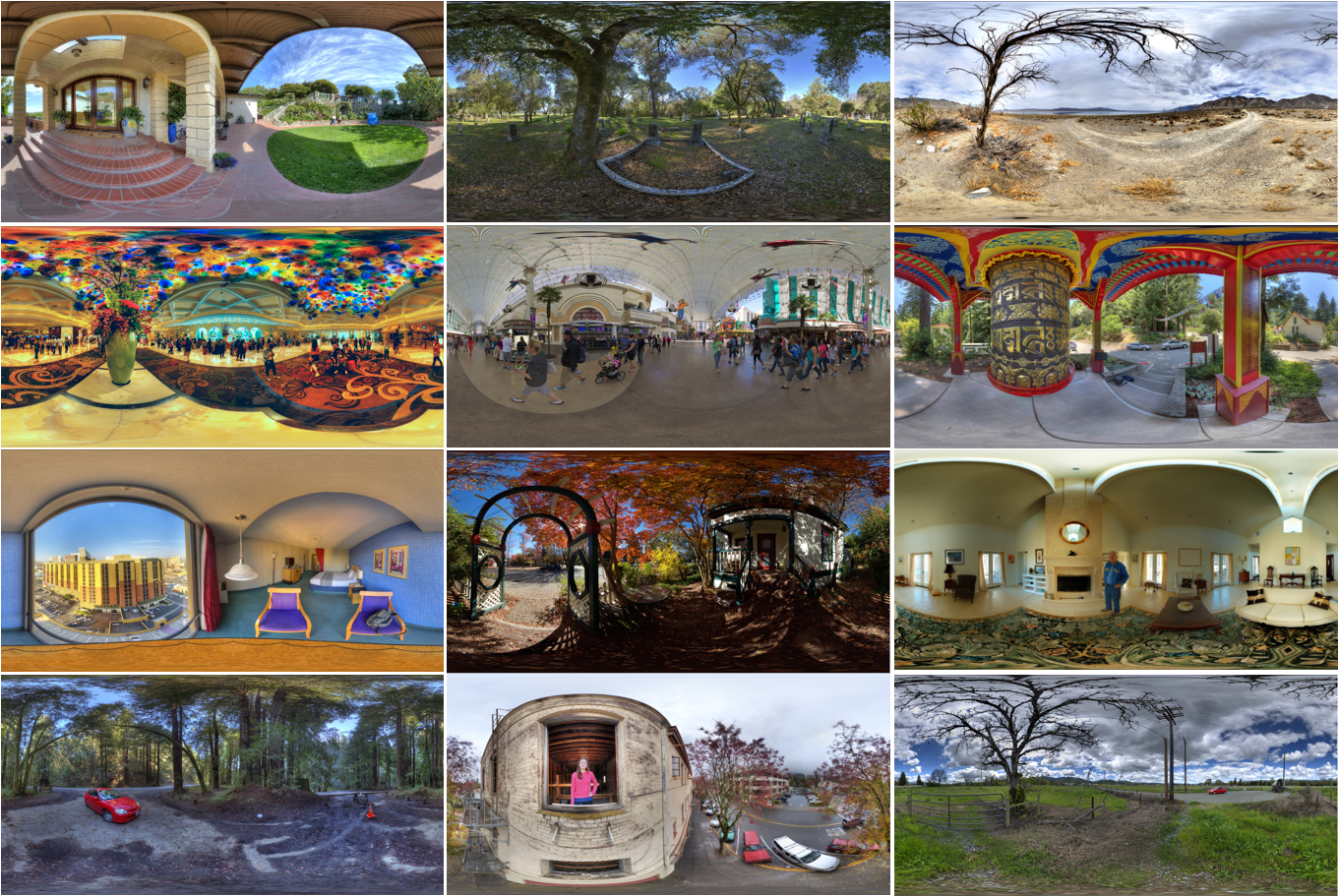}
  \vspace{-0.2cm}
  \caption{The sample of source images in the database.}
  \label{fig:picture1}
\end{figure}

There are 336 images in total in the database, which contains 16 raw images and 320 distorted images. All of the images are in equirectangular formats whose resolution range from 11332$\times$5666 to 13320$\times$6660. The raw images were captured by professional photographers and available under Creative Commons (CC) copyright. Fig. \ref{fig:picture1}. shows several examples of source images in the database.
All raw images are zoomed in and carefully checked to avoid easily observed artifacts, and all of them have close resolutions and perceptual quality. This procedure can avoid the ``intrinsic artifacts" and reduce the influence of the original content's quality.

\vspace{-0.2cm}
\vspace{-0.1cm}

Four types of distortions are introduced in this paper, including JPEG compression, JPEG2000 compression, Gaussian blur and white Gaussian noise, with five distortion levels for each type.
JPEG and JPEG2000 are the two most commonly used compression methods, thus these two methods are selected to simulate the artifacts introduced during compression. Five compression levels are manually set to cover a wide perceptual quality range. All images degraded by these two distortions are compressed in the equirectangular format directly since a lot of omnidirectional content is created and stored in the equirectangular format and then compressed and transmitted.

\begin{figure}[ht]
  \centering
    \includegraphics[width=0.35\textwidth]{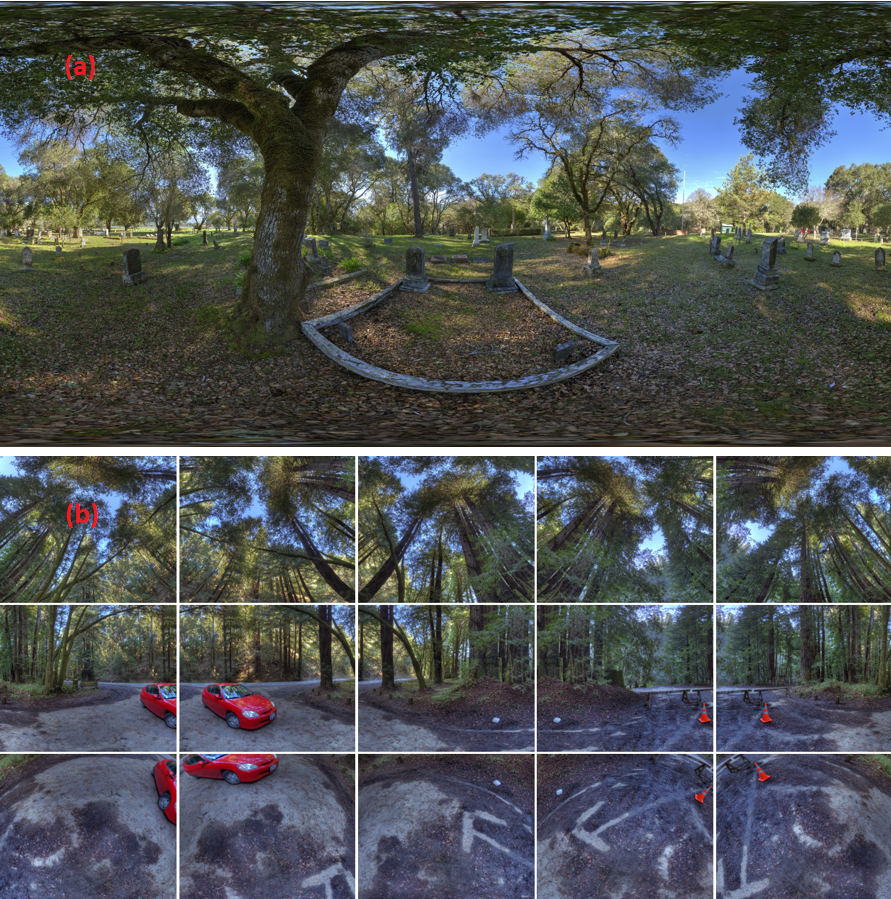}
    \vspace{-0.2cm}
  \caption{One source image and 15 split images. (a) The source image. (b) 15 split images.}
  \label{fig:picture2}
\end{figure}

Gaussian blur and white Gaussian noise are another two commonly encountered distortion types. Considering that the transmission techniques are becoming better and distortions such as blur and noise are less introduced during transmission, we mainly consider the blur and noise introduced during capturing. Because omnidirectional images are usually captured by various kinds of shooting equipment and then stitched, it is difficult to simulate this kind of distortion.
This paper simulates the situation of capturing a series of images using camera array first and then stitching the images into an omnidirectional image. Specifically, we split one source image into 15 small blocks in which the image content is less distorted especially at the south and north poles. As shown in Fig. \ref{fig:picture2}., (a) represents the source image and (b) shows the 15 split images which represent the scenes captured by each camera of the camera array. Since the distortions are generally introduced at the sensor of each camera, Gaussian blur and white Gaussian noise are added to 15 small blocks respectively and then they are stitched back to one omnidirectional image. Following these procedures, the blur and noise are more uniformly added to the image compared with distorting the equirectangular image directly. Similar to the compression distortions, 5 levels of blur and noise distortions are introduced to generate distorted images with varying perceptual quality. As shown in Fig. \ref{fig:picture2}.(b), the split images are overlapped to simulate the realistic image stitching situation.



\vspace{-0.3cm}
\vspace{-0.6cm}

\subsection{Subjective Experiment Methodology}
There are several subjective assessment methodologies recommended by the ITU-R BT.500-11 \cite{ITU}, for instance, single-stimulus (SS), double-stimulus impairment scale
(DSIS) and paired comparison (PC).
Since the subjective experiments have to be conducted in the HMD, and subjects could only view one image at one time, the SS method is chosen in our test.
A 10-point numerical categorical rating method is used to facilitate the rating in HMD.
The HMD we chose is HTC VIVE on account of its excellent graphics display technology and high-precision tracking ability. Since we also need to track the eye movement data, aGlass \cite{aGlass} is used in our tests. aGlass is an excellent VR eye-tracking equipment for HTC VIVE with an error less than 0.5$^{o}$.
Besides that, the head movement data is also tracked by the HTC VIVE. So with the help of HTC VIVE and aGlass, we conduct the experiment to obtain the images' subjective quality scores , head and eye movement data at the same time.

Twenty subjects participate in our experiment. The observers are seated in a swivel chair to ensure they could visualize the whole omnidirectional image.
We also design an interactive system using Unity3D to display the test images and collect the data. At the beginning of the experiment, the subject is asked to calibrate the eye tracking module. Then, we collect the visual attention data for the 16 raw omnidirectional images which are displayed in a sequence. Each image is shown for 20 seconds and there is a 5 seconds' rest time between two images. All subjects are asked to look around in this step to get natural-viewing visual attention data. Next, we conduct a training test to make the subjects familiar with the distortions of the OIQA database. Finally, we conduct the formal quality rating experiment with all images shown in a random order. The eye tracking data is also recorded when conducting the formal experiment. During the experiment, subjects have enough rest time every 10 minutes to avoid fatigue.

\subsection{Data Processing and Analysis}
Through the subjective experiment, three kinds of data are collected: raw subjective quality scores of all images, head and eye movement data. In this section, we will discuss the processing and analyzing of these three kinds of data.

\subsubsection{Subjective Quality Score Processing and Analysis}
Using the subjective quality scores we collected, the MOS of each image can be computed easily as follows:
\begin{eqnarray}
MOS_{j} = \frac{\sum^{N}_{i=1}m_{ij}}{N},
\end{eqnarray}
where N is the number of subjects and $m_{ij}$ is the score assigned by subject i to image j. Sometimes, subjects will give a score which is far away from the mean value. So we use 3$\sigma$ principle to remove these outliers.

\vspace{-0.4cm}
\begin{figure}[ht]
  \centering
  \includegraphics[width=0.35\textwidth]{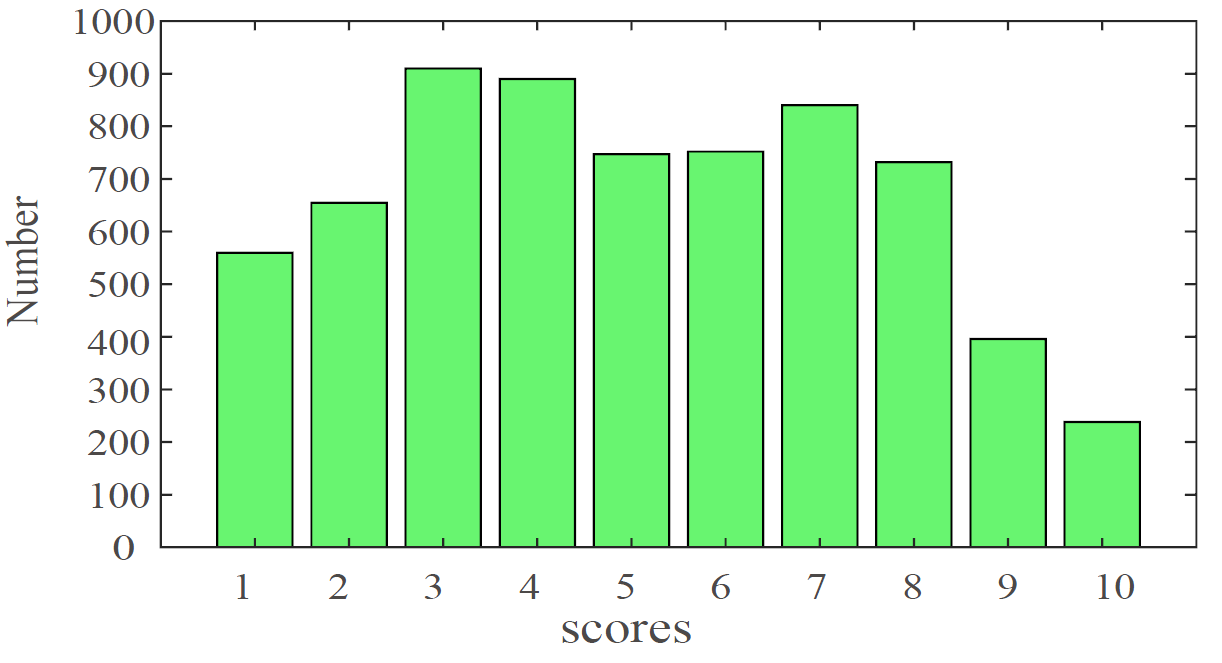}
  \vspace{-0.4cm}
  \caption{Histogram of the subjective quality scores.}
  \label{fig:score}
\end{figure}
\vspace{-0.3cm}

The histogram of the subjective quality scores is illustrated in Fig. \ref{fig:score}.
It is obvious that the scores are distributed across the all perceptual quality range. On account of the high quality and resolution of source images, there are many images whose the subjective quality scores could be up to 9 or 10 points. The subjective quality scores will be given in our database and released.

\subsubsection{Visual Attention Data Processing and Analysis}
We collect the head movement and eye-tracking data of 16 raw images within 20 seconds. We design a method in Unity3D to filter out the saccades in the gaze data to make the visual attention data more reliable. The view direction data collected by the HMD and the eye-tracking data collected by the eye-tracker has been projected from the 3D space to the 2D equirectangular image in our database. We create head-only (view direction centered) saliency maps and head-eye saliency maps using view direction information and eye-tracking data respectively. Specifically, through the procedures above, we get the view-direction positions and eye-fixations of all subjects. All subjects' data for the same image is overlaid to a view-direction position map and an eye-fixation map. We follow the method in \cite{Rai2017A} to get the head-only saliency map and head-eye saliency map using the view-direction position map and eye-fixation map, respectively. The Gaussian filter of 3.34$^{o}$ of visual angle \cite{Curcio1990Human}\cite{Engelke2015Perceived} is applied to fixation maps within the view-port image because it is the true image viewed by subjects. Then the view-port images with spread fixations are back-projected into the sphere-map and then to the final equirectangular visual attention map. The view-direction position, eye-fixation maps and head, head-eye saliency maps will be included in the OIQA database and released.

\begin{figure}[ht]
  \centering
  \includegraphics[width=0.3\textwidth]{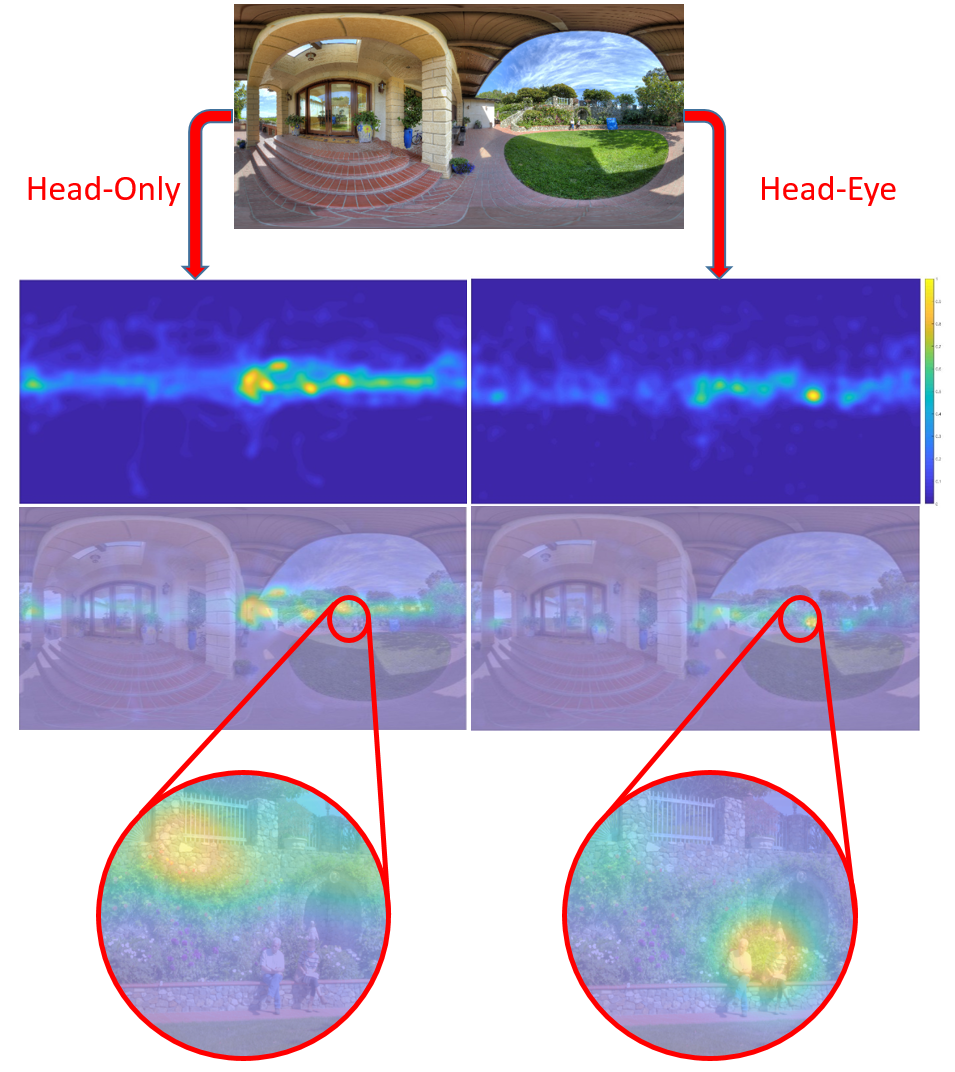}
  \caption{The head-only saliency map and head-eye saliency map.}
  \vspace{-0.4cm}
  \label{fig:saliency}
\end{figure}

Fig. \ref{fig:saliency}.
shows the head-only saliency map and head-eye saliency map of an image. From two saliency maps, it is obvious that the salient regions mainly centralize in the middle of the equirectangular image which is nearby the equator. It is reasonable since when viewing omnidirectional images, the top and bottom region of an equirectangular image are less viewed by the human subjects. Moreover, when viewing omnidirectional images in HMD, we can only see a small part of the whole scene. On account of these two reasons, it is reasonable to assess the quality of an omnidirectional image using visual saliency. As a consequence, we establish the OIQA database which has both subjective quality scores and saliency data. Besides the visual attention data of the 16 original images, we also record the visual attention data during the formal quality rating experiment for researchers in need.

\vspace{-0.4cm}

As shown in Fig. \ref{fig:saliency}, obviously, the head-only saliency map and head-eye saliency map is similar from an overall perspective, but in details, they are quite different. As shown in the zoomed-in figure, the left figure highlights the top-left corner while the right figure highlights the bottom-right corner where there are two peoples. The head-only saliency map reveals the view direction information and the head-eye saliency map reveals the human eye gaze information. Note that, whether for 2D or 3D images, there are some small-size but important or salient regions which provide key information of an image, such as human faces. So, it is reasonable and also important to consider the saliency information in image quality assessment whether it is in two-dimensional or three-dimensional space. In VR HMDs, because of the immersive perception and projection method, the saliency is even more important for omnidirectional IQA compared with traditional IQA. The constructed OIQA database provides such kind of key information to facilitate following research.

\section{Comparison of objective quality assessment on the OIQA Database}

\subsection{Experimental Protocol}
After the experiment, we test nine state-of-the-art full-reference objective IQA models on our database, including FSIM \cite{Zhang2011FSIM}, GMSD \cite{Xue2013Gradient}, GSI \cite{Liu2012Image}, IW-SSIM \cite{Wang2011Information}, MS-SSIM \cite{Wang2004Multiscale}, PSNR, SSIM \cite{Wang2004Image}, VIF \cite{Sheikh2006Image} and VSI \cite{Zhang2014VSI}. When calculating performance, we firstly mapped the predictions of the IQA models to subjective quality ratings through a five-parameter logistic function \cite{min2017blind}\cite{min2017saliency}\cite{min2017unified}:
\vspace{-0.2cm}
\begin{eqnarray}
f(x) = \beta_{1}(\frac{1}{2}-\frac{1}{1+e^{\beta_{2}(x-\beta_{3})}})+\beta_{4}x+\beta_{5},
\end{eqnarray}
in which $x$ denotes the predicted scores; $f(x)$ represents the corresponding mapped score; $\beta_{i} (i = 1, 2, 3, 4, 5)$ are the parameters to be fitted. Then the mapped scores are compared with the subjective scores to measure the performance of the IQA models. Three evaluation criteria are used in this paper are Pearsons linear Correlation Coefficient (PLCC), Spearman rank correlation coefficient (SRCC) and Root mean square error (RMSE). The performance of the nine IQA models is listed in Table \ref{tab:performance}.

\vspace{-0.4cm}
\begin{table}[!htb]
\begin{center}
\caption{Performance of IQA models in terms of PLCC, SRCC and RMSE. The best three performing metrics are highlighted with bold font.} \label{tab:performance}
\vspace{-0.2cm}
\begin{tabular}{c|c c c}
  \hline
  Method & PLCC & SRCC & RMSE\\
  \hline
  \textbf{FSIM} & \textbf{0.9171} & \textbf{0.9110} & \textbf{0.8221}\\
  GMSD & 0.7434 & 0.7388 & 1.3799\\
  \textbf{GSI} & \textbf{0.8992} & \textbf{0.8901} & \textbf{0.9024}\\
  IW-SSIM & 0.7854 & 0.7779 & 1.2767\\
  MS-SSIM & 0.6774 & 0.6664 & 1.5175\\
  PSNR & 0.5088 & 0.4984 & 1.7758\\
  SSIM & 0.5316 & 0.3501 & 1.7473\\
  VIF & 0.7915 & 0.7876 & 1.2607\\
  \textbf{VSI} & \textbf{0.9060} & \textbf{0.9020} & \textbf{0.8729}\\
  \hline
\end{tabular}
\end{center}
\end{table}
\vspace{-0.7cm}

\subsection{Performance Comparison}
As seen in Table \ref{tab:performance}, The three best-performing metrics are FSIM, GSI and VSI and the performance is fairly good. These three IQA models also perform pretty well on tractional IQA databases, which suggests that omnidirectional IQA and traditional IQA have a lot in common. But we still believe there is a certain room for improvement, such as using saliency to promote omnidirectional IQA. Except for these three models, other state-of-the-art IQA models perform not well, and they undergo some performance drop when transferring from traditional images to omnidirectional images. There is also a lot of room to improve these models.

\subsection{Differences between Omnidirectional IQA and Traditional IQA}
\begin{figure}[!htb]
  \centering
  \includegraphics[width=0.48\textwidth]{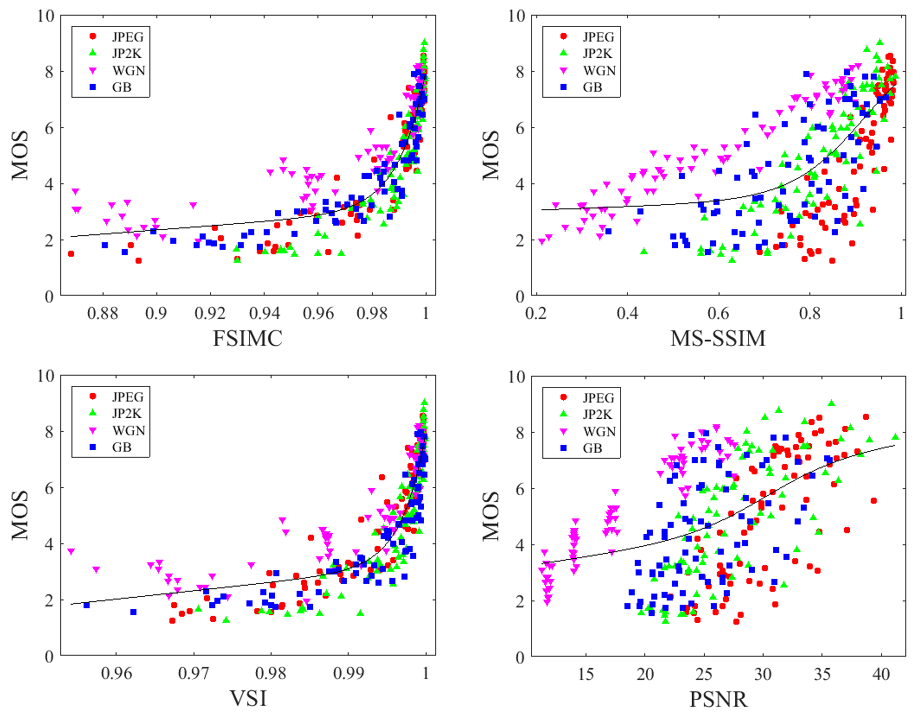}
  \vspace{-0.4cm}
  \caption{Scatter plots of four representative IQA models, including FSIMc, VSI, MS-SSIM, PSNR. JPEG: JPEG compression; JP2K: JPEG2000 compression; WGN: white Gaussian noise; GB: Gaussian blur.}
  \label{fig:scatter}
\end{figure}

We select four representative IQA models, including FSIMc, MS-SSIM, VSI, and PSNR, and illustrate their scatter plots in Fig. 5. FSIMc and VSI are selected because of their high performance, whereas MS-SSIM and PSNR are two classical IQA models.
As shown in Fig. \ref{fig:scatter}, the scatter points of the distortion type white Gaussian noise (WGN), whose the color is magenta, are always far away from the fitted curve compared with other distortion types and they are almost always higher than other scatter points of other distortion types. It means that all these four IQA models have predicted lower quality scores than the ideal values for distortion type WGN.
This phenomenon is observed not only in these four representative IQA models, but also in various IQA models, thus we believe it is caused by the subjective ratings rather than the objective IQA models. Moreover, most IQA models show quite consistent predictions for all kinds of distortions in traditional images.
We suggest that the exceptional subjective ratings is caused by a human preference of high frequency content when viewing VR stimuli. With such kind of high frequency content, the observers can have a more comfortable visual experience. It is also caused by the limited displaying effects of current HMD, and subjects can not see so many details of the content compared with tractional displays. Subjects are annoyed with the losing of image details. Among all four types of distortions, Gaussian noise adds some high frequency information to the image, while the rest three distortions reduce the high frequency information and image details. Some following work can be done considering this phenomenon. It is also very important to incorporate visual saliency into omnidirectional IQA, since visual saliency always highlights the high frequency content of the image.

\vspace{-0.4cm}
\vspace{-0.4cm}

\section{Conclusion and Future Work}
This paper has investigated an emerging quality assessment problem of omnidirectional images in the VR environment. We first construct an omnidirectional image quality assessment (OIQA) database, including 16 source images and 320 degraded images distorted by four most commonly encountered distortions. We collect the subjective quality scores, view direction information and eye-tracking data during the experiment and all of the data will be included in the OIQA database and released. By comparing nine objective IQA models on the OIQA database, we suggest that humans prefer high frequency content and image details in VR HMDs, and the losing of image details can do a lot of harm to the visual experience in the VR case. Some following work can be done to correct such deviation when applying traditional IQA models to omnidirectional IQA. Visual saliency can be also utilized for omnidirectional IQA, and we believe that visual saliency can promote the performance considering the extreme importance of visual attention in the VR environment.



%
%



%

%
%

\bibliographystyle{IEEEtran}
\bibliography{bare_conf}

\end{document}